\documentclass[runningheads]{llncs}
\usepackage{graphicx}
\usepackage{amsmath,amssymb} 
\usepackage{tabularx}
\usepackage{multirow}
\usepackage{subfigure}
\usepackage{dsfont}
\usepackage{algorithm}
\usepackage{colortbl}
\usepackage{booktabs}
\usepackage{tabularx}
\usepackage[table,xcdraw]{xcolor}
\usepackage{algorithm}
\usepackage[]{algpseudocode}
\usepackage{color}\usepackage[width=122mm,left=12mm,paperwidth=146mm,height=193mm,top=12mm,paperheight=217mm]{geometry}

\newtheorem{prop}{Proposition}

\newcommand*{\QEDB}{\hfill\ensuremath{\blacksquare}}%

\makeatletter
\def\hlinewd#1{%
	\noalign{\ifnum0=`}\fi\hrule \@height #1 \futurelet
	\reserved@a\@xhline}

\begin{document}
\pagestyle{headings}
\mainmatter
\title{Efficient Splitting-based Method for Global Image Smoothing} 

\titlerunning{Efficient Splitting-based Method for Global Image Smoothing}

\authorrunning{Y. Kim et al.}

\author{Youngjung Kim$^{1}$, Dongbo Min$^2$, Bumsub Ham$^3$, and Kwanghoon Sohn$^1$}
\institute{$^1$Yonsei University, $^2$Chungnam National University, $^3$Inria}

\maketitle

\begin{abstract}
Edge-preserving smoothing (EPS) can be formulated as minimizing an objective function that consists of data and prior terms.
This global EPS approach shows better smoothing performance than a local one that typically has a form of weighted averaging, at the price of high computational cost.
In this paper, we introduce a highly efficient splitting-based method for global EPS that minimizes the objective function of ${l_2}$ data and prior terms (possibly non-smooth and non-convex) in linear time.
Different from previous splitting-based methods that require solving a large linear system, our approach solves an equivalent constrained optimization problem, resulting in a sequence of 1D sub-problems.
This enables linear time solvers for weighted-least squares and -total variation problems.
Our solver converges quickly, and its runtime is even comparable to state-of-the-art local EPS approaches.
We also propose a family of fast iteratively re-weighted algorithms using a non-convex prior term.
Experimental results demonstrate the effectiveness and flexibility of our approach in a range of computer vision and image processing tasks.
\end{abstract}

\section{Introduction}
Edge-preserving smoothing (EPS) has attracted a strong interest in the fields of image processing and computer vision.
Predominantly, it appears in a manipulation task that requires decomposing an image into a piecewise smooth layer and a detail layer.
These layered signals can be recombined to match various application goals, e.g., detail enhancement and tone mapping \cite{Farbman2008}.
Recent works on joint smoothing provide a new paradigm, enabling various applications such as dense correspondence \cite{Rhemann2011}, joint upsampling \cite{Kopf2007,Park2011}, and texture removal \cite{Xu2012}.
The problem of segmentation \cite{Kim2008} and visual saliency \cite{Perazzi2012} may also be interpreted as a joint smoothing problem.
The basic idea of joint smoothing is to provide structural guidance of how the smoothing should be performed.
Thus, it is assumed that the guidance signal has enough information to alter structures in the input image.
In this context, global optimization-based methods \cite{Farbman2008,Bi2015,Ham2015} are advocated.
They find an optimal solution to an objective function that consists of a data fidelity term and a prior term.
Thanks to such global formulation, the optimization-based methods show the state-of-the-art performance compared with local EPS approaches that typically have a form of weighted averaging \cite{Farbman2008}.
However, this outperformance can be achieved only at the price of high computational cost, mainly arising from solving the global objective function.
The optimization-based methods are still an order of magnitude slower than local ones \cite{He2013,Gastal2011,Chen2007}, even with recent acceleration techniques \cite{Krishnan2013,Wang2008,Afonso2010}.
Progress in hardware will make an efficient implementation possible, but it is not unlikely that an image resolution will increase as well.
Accordingly, a different optimization technique is needed for a highly efficient global EPS methods.

In this paper, we formulate a global EPS (e.g., based on weighted-least squares or weighted-total variation) as an equivalent constrained optimization problem.
This formulation results in a sequence of sub-problems that is much easier to optimize.
The computational efficiency of our approach is due to a kind of variable splitting techniques.
However, unlike the previous splitting-based methods\footnote{The most expensive operation in previous methods, in the senses of alternating minimization \cite{Wang2008,Afonso2010} or half quadratic optimization \cite{Xu2011}, is fast Fourier transforms with $O(n\log n)$ complexity.} \cite{Wang2008,Afonso2010},
our formulation enables using a highly efficient, linear time algorithm in both weighted-least squares (WLS) and -total variation (WTV) problems.
As a result our approach has a time complexity linear to the number of image pixels.
Another appealing aspect is that it converges with only few iterations.
We also propose fast iteratively re-weighted algorithms for an objective function using a non-convex prior term.
Note that the previous splitting-based methods in \cite{Wang2008,Afonso2010} are not directly applicable to many non-convex priors of practical relevance \cite{Schmidt2014}.

\section{Problem formulation and Analysis}
EPS can be formulated as finding a solution of an objective function.
We can find the solution by using popular iterative methods \cite{Saad2003} or splitting-based approaches \cite{Wang2008,Afonso2010},
depending on the prior terms used in the objective function.
We start with a basic formulation for EPS to provide some intuition.
In the following, the subscript $p$ denotes the 2D spatial location of a pixel.

Given an input image $f$ and a guidance image $g$, a desired output
$u$ is obtained by minimizing the following objective function:

\begin{equation}
E\left( u \right) = \sum\limits_p {\left( {(u - f)_p^2 + \lambda \sum\limits_{j \in \{ 1,2\} } {{w_{j,p}}\phi {{({D_j}u)}_p}} } \right)},
\end{equation}
where ${D_1}$ and ${D_2}$ are discrete implementations of the derivative in horizontal and
vertical directions,  respectively.
$\lambda$ controls the strength of the smoothing.
$g$ can be the input image $f$ or a different signal correlated with $f$.
The weight $w_{j,p}=\exp ( - ({D_j}g)_p^2/\kappa )$ is defined using $g$ and constant $\kappa$.
The potential function $\phi$ and the weight function $w$ allow one to
employ various image priors that behave differently in preserving
or smoothing image features.
We always assume that $f$ and $g$ have the same width ($W$) and the height ($H$).

\subsection{WLS for EPS}
When $\phi  = {\tau^2}$, the objective function of (1) becomes quadratic, and it corresponds to the WLS framework \cite{Farbman2008}.
The minimizer $u$ satisfies the following linear system:

\begin{equation}
\bigg( {{\bf{I}} + \lambda \sum\limits_{j \in \{ 1,2\} } {{\bf{D}}_j^T{{\bf{W}}_j}{\bf{D}}_j^{}} }\bigg){\bf{u}} = {\bf{f}},
\end{equation}
where ${{\bf{D}}_j}$ is a discrete difference matrix, ${{\bf{W}}_j}$ is a diagonal matrix containing the weights ${w_j}$, and ${\bf{I}}$ is an identity matrix.
Iterative solvers such as Jacobi and conjugate gradient methods \cite{Saad2003} are
applicable to solve the sparse linear system of (2).
Since these methods consist of matrix-vector multiplications, each iteration runs in linear time to an image size.
However, the number of iterations, required for achieving a particular accuracy, depends on the matrix dimension ($HW\times HW$) \cite{Krishnan2013}, and hence the
computational cost is considerable.
Despite recent progress in preconditioning techniques, all the existing
solvers are still an order of magnitude slower than the state-of-the
art local EPS approaches \cite{He2013,Gastal2011}.
Moreover, the cost of constructing the preconditioner may outweigh
the speed gain from the improved conditioning.

Similarly, the preconditioning techniques \cite{Krishnan2013} may not
accelerate EPS algorithms using the iteratively
re-weighted least squares (IRLS) method \cite{Chartrand2008}. An intermediate linear
system of the IRLS method varies during \emph{external} iterations
and thus a series of preconditioners should be constructed, leading
to a huge amount of computational overhead.

\subsection{WTV for EPS}
The WTV prior, $\phi = |\tau|$, often achieves a better capability along boundaries \cite{Xu2012,Bi2015}, but it needs more computational cost than using $\phi  = {\tau^2}$ in the WLS.
A common approach to minimizing the
WTV objective function is to exploit the variable-splitting
and penalty techniques, as follows \cite{Wang2008}:

\begin{equation}
E\left( {u,v,\beta } \right) = \sum\limits_p {\left( {(u - f)_p^2 + \sum\limits_{j \in \{ 1,2\} } {\left( {\frac{\beta }{2}({v_j} - {D_j}u)_p^2 + \lambda {w_{j,p}}{{\left| {{v_j}} \right|}_p}} \right)} } \right)} ,
\end{equation}
where $\beta$ is a penalty parameter.
An auxiliary variables $v_1$ and $v_2$ are introduced for
an alternative minimization of the data and the prior terms.
When $v=(v_1,v_2)$ is fixed, minimizing the objective function of (3) with respect to ${u}$ can be solved in a closed-form, and vice versa.
The solver is hence iteratively applied while updating $u$, $v$, and $\beta$:

\begin{equation}
\begin{array}{l}
{u^{t + 1}} = \mathop {\arg \min }\limits_u E(u,{v^t},{\beta ^t}),\\
{v^{t + 1}} = \mathop {\arg \min }\limits_v E({u^{t + 1}},v,{\beta ^t}),\\
{\beta ^{t + 1}} = \alpha {\beta ^t},
\end{array}
\end{equation}
where $\alpha {\rm{ > 1}}$ is a continuation parameter.
Since ${v^{t+1}}$ can be obtained by \emph{soft-thresholding} \cite{Wang2008,Afonso2010,Bi2015},
the computational cost primarily lies in the $u$-subproblem:
\vspace{5pt}
\begin{equation}
\bigg( {{\bf{I}} + \frac{\beta }{2}\sum\limits_{j \in \{ 1,2\} } {{\bf{D}}_j^T{\bf{D}}_j^{}} } \bigg){{\bf{u}}^{t + 1}}
= \frac{\beta }{2}\sum\limits_{j \in \{ 1,2\} } {{\bf{D}}_j^T{\bf{v}}_j^t}  + {\bf{f}}.
\end{equation}
Although numerous methods, such as alternating direction method of multipliers (ADMM) \cite{Afonso2010} and split-Bregamn (SB) \cite{Bi2015}, have been proposed,
they use the fast Fourier Transform\footnote{The linear system (5) can be diagonalized by FFT. Thus, solving (5) requires three FFT calls.} (FFT) to update the variable $u$.
As a result, the computational cost required for solving (3) is $O(n\log n)$ per iteration ($n=HW$).

\section{Method}
In this section, we first propose an efficient method of
minimizing (1) when $\phi  = \tau^2$ or $|\tau|$. We then apply it to solve non-convex objective functions.
The key idea is to decompose (1) into each
spatial dimension with a proper variable splitting, and then to use a constrained optimization technique.

Consider the following optimization problem with linear equality constraint:

\begin{equation}
\mathop {\min }\limits_{\scriptstyle\;\;\;{\kern 1pt} u,v\hfill\atop
\scriptstyle\,{\rm{s}}{\rm{.t}}{\rm{. u = v}}\hfill} \sum\limits_p {\left( {\sum\limits_{o \in \{ u,v\} } {\frac{1}{2}(o - f)_p^2 + }
\lambda \left( {{w_{1,p}}\phi {{({D_1}u)}_p} + {w_{2,p}}\phi {{({D_2}v)}_p}} \right)} \right),}
\end{equation}
We again denote by $v$ an auxiliary variable. In our formulation, the size of $v$ is equal to the input $f$.
Then, it is clear that (6) is equivalent to the original problem (1), under the constraint $(u=v)$.
The penalty decomposition method\footnote{We have tested the
augmented Lagrangian method \cite{Nocedal2006} to avoid using large $\beta$ values
(continuation), but found this method did not improve a
convergence rate, while increasing a memory demand (due to Lagrange
multiplier).} \cite{Nocedal2006}, associated with (6) can be written as:

\begin{equation}
\mathop {\min }\limits_{u,v} \sum\limits_p {\left( {\sum\limits_{o \in \{ u,v\} }
{\frac{1}{2}(o - f)_p^2 + \frac{\beta }{2}(u - v)_p^2 + }
\lambda \left( {{w_{1,p}}\phi {{({D_1}u)}_p} + {w_{2,p}}\phi {{({D_2}v)}_p}} \right)} \right).}
\end{equation}
This problem can be solved with block coordinate descent by minimizing (7) with respects to $u$ and $v$, alternately
\footnote{For scalar variables, $\mathop {\arg \min }\limits_e a{(e - c)^2} + b(e - d)^2 = \mathop {\arg \min }\limits_e (a + b){(e - \frac{{ac + bd}}{{a + b}})^2}$.}.

\begin{equation}
{u^{t + 1}} = \mathop {\arg \min }\limits_u \sum\limits_p {\left( {(u - \tilde f)_p^2 + \frac{2\lambda }{{1 + {\beta ^t}}}{w_{1,p}}\phi {{({D_1}u)}_p}} \right),}
\end{equation}

\begin{equation}
{v^{t + 1}} = \mathop {\arg \min }\limits_v \sum\limits_p {\left( {(v - \bar f)_p^2 + \frac{2\lambda }{{1 + {\beta ^t}}}{w_{2,p}}\phi {{({D_2}v)}_p}} \right),}
\end{equation}
where $\tilde f = {(1 + {\beta ^t})^{ - 1}}(f + {\beta ^t}{v^t})$,
${\rm{ }}\bar f = {(1 + {\beta ^t})^{ - 1}}(f + {\beta ^t}{u^{t +1}})$,
and ${\beta ^{t + 1}} = \alpha {\beta ^t}$.

Now, we are ready to show the advantage of our formulation.
As $D_1$ represents a difference operator with respect to the horizontal axis,
we can decompose (8) into sub-problems defined with 1D horizontal signals only.
By introducing a 1D slack variable $z$, we have:

\begin{equation}
{u^{t + 1,h}} = \mathop {\arg \min }\limits_z \sum\limits_x {\left( {(z - {{\tilde f}^h})_x^2 + \frac{{2\lambda }}{{1 + {\beta ^t}}}w_{1,x}^h\phi {{({D_1}z)}_x}} \right),}
\end{equation}
The super-script $h$ denotes a horizontal signal along the $x$
dimension ($x = 1, \ldots ,{W}$). This 1D optimization process
is repeated for all horizontal signals ($H$ in number). Note that a similar result can be obtained for the auxiliary variable $v$. In this case, (9) is
decomposed into a sub-problem with a 1D vertical signal along the
$y$ dimension ($y = 1, \ldots ,{H}$). \QEDB

To the best of our knowledge, the variable splitting technique has
been applied in such a way that
${\left[ {\begin{array}{*{20}{c}}
{{D_1}u}, {{D_2}u}
\end{array}} \right]} = [v_1,v_2]$
, yielding $O(n\log n)$ complexity algorithm $(n=HW)$, as explained in Section 2.
The previous variable splitting \cite{Afonso2010,Bi2015,Xu2011} aims to transfer $D_1u$ and $D_2u$ out of the potential function $\phi$ by introducing the auxiliary variable $v_1$ and $v_2$, respectively.
In contrast, we utilize it to decompose the original problem of (1) into a series of 1D sub-problems.
This not only leads to easily solvable sub-problems, but also significantly improves the convergence rate of the algorithm (see Section 4).
In the following, we present an efficient method of solving the objective form of (10) defined
with a 1D horizontal signal. Its vertical counterpart can be
optimized in the same manner.

\subsection{1D Fast solver}

\subsubsection{WLS}
With $\phi=\tau^2$, (10) can be rewritten with a 1D horizontal signal
${\tilde f^h}$ and a guide signal $g^h$ as

\begin{equation}
\mathop {\arg\min }\limits_z \sum\limits_x {\left( {(z - {{\tilde f}^h})_x^2 + \tilde w_{1,x}^h({D_1}z)_x^2} \right)} ,
\end{equation}
where $\tilde w_{1,x}^h = {(1 + {\beta ^t})^{ - 1}}(2\lambda w_{1,x}^h)$.
The 1D output $z$ that minimizes the above equation is obtained by solving the following linear system of size $W \times W$.

\begin{equation}
\left( {{\bf{I}} + {\bf{D}}_1^T{\bf{\tilde W}}_1^h{\bf{D}}_1^{}} \right){{\bf{z}}^h} = {{\bf{\tilde f}}^h}.
\end{equation}
Note that the size of ${{\bf{D}}_1^{}}$ and ${\bf{I}}$ is $W \times W$, not $HW \times HW$ as in (2).
Interestingly, the problem (12) becomes much easier to solve than (2) since
$\tilde L$ is a tridiagonal matrix.
We can solve this equation with $O(n)$ cost ($n=W$) by the Thomas algorithm \cite{Golub1996}.
It consists of forward-backward steps. More details can be found in the supplementary material.

\subsubsection{WTV}
When $\phi = |\tau|$, (10) is written as follows:

\begin{equation}
\mathop {\arg\min }\limits_z \sum\limits_x {\left( {(z - {{\tilde f}^h})_x^2 + \tilde w_{1,x}^h{{\left| {{D_1}z} \right|}_x}} \right)} ,
\end{equation}
The IRLS algorithm can be applied to solve (13) approximately \cite{Chartrand2008}.
However, there exists a non-iterative, $O(n)$ method for the 1D total variation \cite{Condat2013}.
While this method is designed to solve 1D (un-weighted) total variation,
it is possible to extend it to minimize 1D WTV.
Note that this extension enables the fast iteratively re-weighted L1 (IRL1) algorithm for 2D image smoothing. See Section 3.3 for details.

We introduce the (Fenchel-Moreau) dual form of (13) as follows:

\begin{equation}
\mathop {\min }\limits_s \sum\limits_x {({{\tilde f}^h} - D_1^Ts)_x^2} ,\;\; {\rm{ s}}{\rm{.t}}{\rm{. }}\;\;{\left| s \right|_x} \le {\tilde w}_{1,x}^h, \;\;{\rm{ }}{s_1} = {s_W} = 0,\;
\end{equation}
where $s$ is the dual variable.
Once the solution ${s^*}$ of the problem (14) is found, we can recover the solution ${z^*}$ of its primal form by

\begin{equation}
z_x^* = \tilde f_x^h - s_x^* + s_{x - 1}^*,\;\;{\rm{ for }}\;\;1 \le x \le W.
\end{equation}
The optimality condition which characterizes the solutions $z^*$ and ${s^*}$ is then expressed as

\begin{equation}
\left\{ {\begin{array}{*{20}{c}}
{s_x^* = \tilde w_{1,x}^h\quad \;\quad \;\;}&{{\rm{if}}\;z_{x + 1}^* > z_x^*}\\
{\begin{array}{*{20}{c}}
{s_x^* =  - \tilde w_{1,x}^h\,\quad \quad }\\
{s_x^* \in [ - \tilde w_{1,x}^h,\tilde w_{1,x}^h]}
\end{array}}&{\begin{array}{*{20}{c}}
{{\rm{if}}\;z_{x + 1}^* < z_x^*}\\
{{\kern 1pt} {\rm{if}}\;z_{x + 1}^* = z_x^*.}
\end{array}}
\end{array}} \right.
\end{equation}
More details about the derivation of (14) and (16) are available at the supplementary material.

\begin{algorithm}
\caption{Fast global image smoothing}\label{euclid}
\begin{algorithmic}[1]
\Procedure{Fast image smoothing using $\phi$}{}
\State Initialize $u^{(t=1)}=v^{(t=1)}=f,$ $\beta^1$, $\alpha$
\For{$t=1:T$}
\For{$y=1:H$}
\State $\tilde f^{h}(x) = {(1 + {\beta ^t})^{ - 1}}(f(x,y) + {\beta ^t}{v^t(x,y)})$ for all $(x = 1, \ldots ,{{W}})$
\State ${\tilde w^h_1}(x) = {(1 + {\beta ^t})^{ - 1}}(2\lambda {w_1}(x,y))$ for all $x$
\State Compute $z$ minimizing (11) or (13) according to $\phi$
\State $u^{t+1}(x,y) = z(x)$ for all $x$
\EndFor
\For{$x=1:W$}
\State $\bar f^{v}(y) = {(1 + {\beta ^t})^{ - 1}}(f(x,y) + {\beta ^t}{u^{t+1}(x,y)})$ for all $(y = 1, \ldots ,{{H}})$
\State ${\tilde w^v_1}(y) = {(1 + {\beta ^t})^{ - 1}}(2\lambda {w_2}(x,y))$ for all $y$
\State Compute $z$ minimizing (11) or (13) according to $\phi$
\State $v^{t+1}(x,y) = z(y)$ for all $y$
\EndFor
\State ${\beta ^{t + 1}} = \alpha {\beta ^t}$
\EndFor
\EndProcedure
\end{algorithmic}
\end{algorithm}

\begin{figure*}[t]
\centering
\renewcommand{\thesubfigure}{}
\subfigure[(a) The WLS smoothing]
{\includegraphics[width=0.49\linewidth]{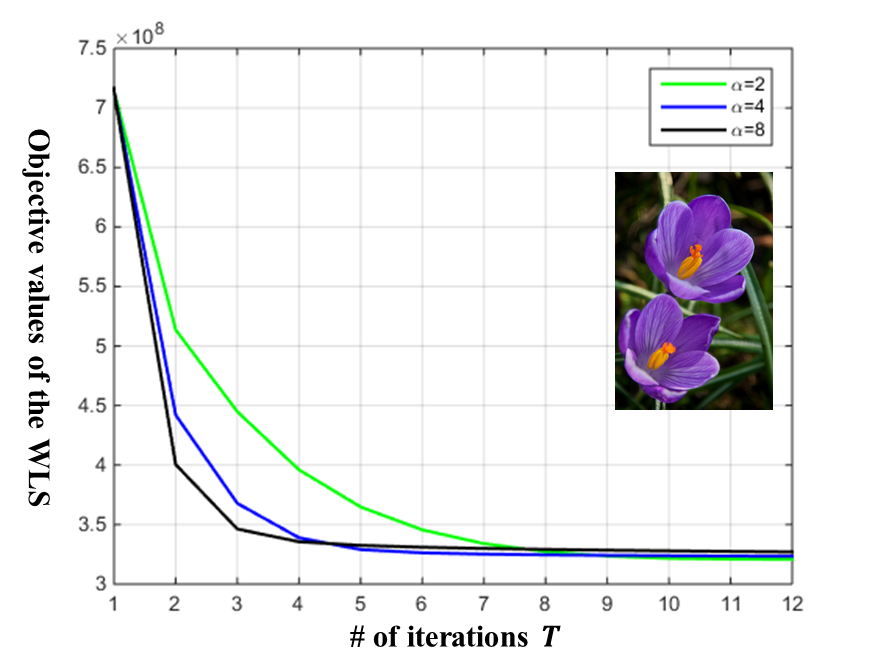}}\hfill
\subfigure[(b) The WTV smoothing]
{\includegraphics[width=0.49\linewidth]{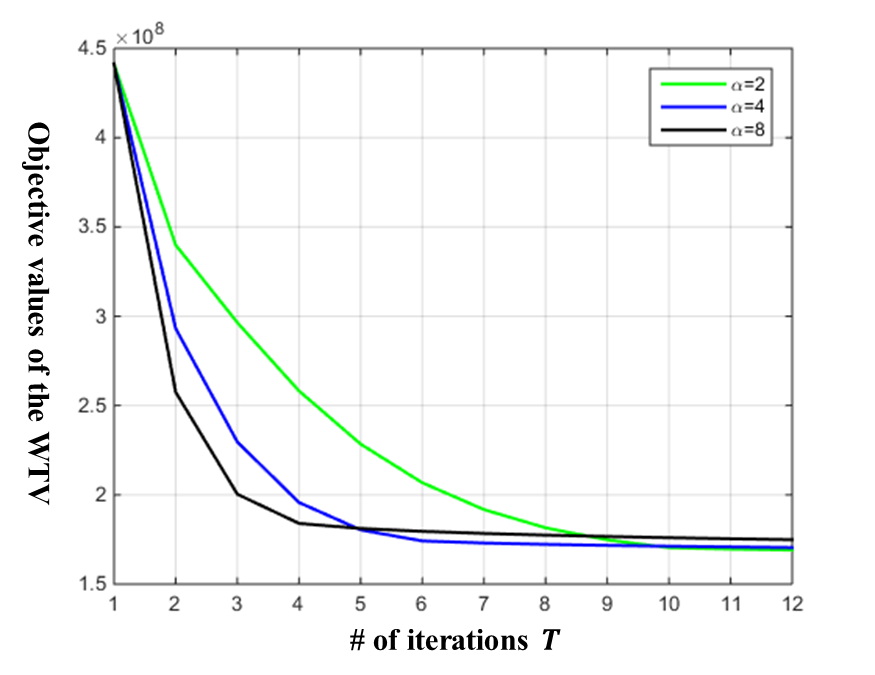}}\hfill
\caption{The energy evolutions of our approach, depending on the continuation parameter $\alpha$. (a) Our fast WLS and (b) Our fast WTV.
$\lambda$ and $\kappa$ are set to $400$ and $7.65$, respectively.}
\label{img:2}
\end{figure*}

\subsection{2D Smoothing algorithm and properties}
The proposed algorithm in the case of 2D image smoothing is summarized
in Algorithm 1. Given an input $f$, a guidance $g$ and a smoothing
parameter $\lambda$, the global 1D smoothing operations are
sequentially performed along with the horizontal and vertical
directions. At each iteration, the input for 1D horizontal smoothing
is re-calculated as the linear combination of the auxiliary variable $v^t$ and $f$ (line 5).
In the same manner, 1D vertical smoothing is applied to the linear combination of $u^{t+1}$ and $f$ (line 11).
These reflect the smoothing effect from the other side.
The 1D smoothing parameter gradually decreases by factor of $2/(1 + {\beta ^t})$ (line 6, 12).
The algorithm is terminated after $t=T$ iterations.

Although the original formulation (1) is decomposed into a series of
sub-problems, we compute exact solutions of the objectives (8) and
(9) defined on the 1D dimension. This property considerably accelerates the convergence speed of the algorithm.
Moreover, since these sub-problems can be decoupled, a straightforward parallelization is possible.

\begin{prop}
As $t \to \infty $, Algorithm 1 is convergent.
\end{prop}
\paragraph*{Proof}
See the supplementary material. \QEDB

\vspace{3pt}
Our approach is not rotationally invariant, as the original
formulation of (1) enforces the smoothness term to be aligned for each axis
individually.
For the WLS smoothing, we do not observe
any visible artifacts in all experiments.
A similar observation can be found in \cite{Farbman2008}.
When $\phi = |\tau|$, it prefers object boundaries which are minimal in the Manhattan ($L_1$) distance.
This may give blocky artifacts in the region boundaries.
In order to alleviate this problem, Algorithm 1 can be customized to perform a smoothing using
8-neighborhood system $\mathcal{N}_8$ $ - $ we should introduce two more
auxiliary variables (by symmetry) and additionally iterate diagonal and anti diagonal passes\footnote{But, all results in
this paper are obtained using a $\mathcal{N}_4$ neighborhood system.}.

\subsubsection{Continuation parameter $\alpha$}
Our approach uses a continuation parameter $\alpha$ to gradually increase a penalty parameter $\beta$, at each iteration.
This ensures that the auxiliary variable $v$ should be close to $u$, and thus Algorithm 1 is convergent (Proposition 1).
Fig. 1 shows the energy evolutions of WLS and WTV with different values of $\alpha$=2, 4, and 8.
Using a large $\alpha$ makes the convergence faster, but the objective value after the convergence becomes slightly higher than that of using a small $\alpha$.
In this paper, we set $\alpha=4$ by considering the trade-off between speed and accuracy.

\subsection{Extension to fast iteratively re-weighted algorithms}
Our approach can be extended for a wider class of potential functions by using iteratively re-weighted algorithms \cite{Ochs2015}.
Let us consider general heavy-tailed potentials $\psi $, then the original objective function of (1) is written as

\begin{equation}
E\left( u \right) = \sum\limits_p {\left( {(u - f)_p^2 + \lambda \sum\limits_{j \in \{ 1,2\} } {{w_{j,p}}\psi {{({D_j}u)}_p}} } \right)}.
\end{equation}
The principle of iteratively re-weighted algorithms is to find a convex function, which serves as the upper bound of (17).
It allows for explicit algorithms according to the construction of this bound, i.e., IRLS vs IRL1.
In the following, we will use $k = \left( {1, \ldots ,K} \right)$ to represent the \emph{external} iteration index used to minimize (17).

\subsubsection{Fast IRLS (FIRLS)}
The IRLS approach uses the fact that many potentials can be seen as minima of the quadratic upper bound \cite{Ochs2015}.
Using an intermediate estimate ${u^{k}}$, we obtain the following update rule:

\begin{equation}
{u^{k + 1}} = \mathop {\arg \min }\limits_u \sum\limits_p {\left( {(u - f)_p^2 + \lambda \sum\limits_{j \in \{ 1,2\} } {{a_{j,p}}({D_j}u)_p^2} } \right)} ,
\end{equation}
where ${a_{j,p}} = {w_{j,p}}\frac{{\psi '{{({D_j}{u^k})}_p}}}{{2{{({D_j}{u^k})}_p}}}$.
Minimizing (18) is equivalent to solving a linear system (2), except that a Laplacian matrix is computed with $a_{j,p}$.
Thus, we can obtain $u^{k+1}$ using our fast WLS solver with $T$ \emph{internal} iterations.
This extension is attractive in several aspects.
First, our approach can be applied to a range of applications, which require for sparse image gradients and sharp edges (See Section 5).
Second, our approach involves only simple arithmetic operations for solving (18).
While most existing linear solvers \cite{Saad2003} suffer from additionally computing the preconditioner at each \emph{external} iteration, as an intermediate linear system varies with $k=(1,...,K)$.

\subsubsection{Fast IRL1 (FIRL1)}
Theoretically, the IRLS algorithm can be applied only when the
potential function is well approximated with a quadratic upper bound
\cite{Ochs2015}.
This, however, does not cover interesting functions
such as $\psi  = \log (1 + |\tau|)$ and ${L_p}$ ($p<1$),
which are concave and non-differentiable at origin.
Here we extend our fast WTV solver to the FIRL1 algorithm:

\begin{equation}
{u^{k + 1}} = \mathop {\arg \min }\limits_u \sum\limits_p {\left( {(u - f)_p^2 + \lambda \sum\limits_{j \in \{ 1,2\} } {{b_{j,p}}\left| {{D_j}u} \right|_p} } \right)} ,
\end{equation}
where ${b_{j,p}} = {w_{j,p}}\partial \psi {({D_j}{u^k})_p}$, $\psi$ is concave, and $\partial$ denotes the sub-gradient.
The algorithm exploits a convex function obtained by linearizing the potential $\psi$.
Note that (19) serves as the upper bound of (17) due to concavity of $\psi$.

Obviously, each of the IRL1 iterations is the WTV problem, which can be solved efficiently using our solver.
The pseudo-code for our fast iteratively re-weighted algorithms is provided in Algorithm 2.

\begin{algorithm}
\caption{Fast iteratively re-weighted algorithms}\label{euclid}
\begin{algorithmic}[1]
\Procedure{Fast image smoothing using $\psi$}{}
\State Initialize $u^{(k=1)}$
\For{$k=1:K$}
\State Construct the sub-problem (18) or (19)
\State Compute $u^{k+1}$ using Algorithm 1
\EndFor
\EndProcedure
\end{algorithmic}
\end{algorithm}

\section{Experimental Validation}
We evaluate the convergence and runtime performance of our solver.
The experiments are simulated with a single Intel i7 3.4GHz CPU. Our approach is easy to implement and we will release the source code at public website.
All compared methods have been implemented in MATLAB with MEX interface. Primary parameters in our approach are set as:
$\beta^1=1$, and $\alpha=4$. At each iteration, we gradually increase $\beta$ by factor of $\alpha$. The smoothing parameters $\lambda$ and $\kappa$ are set to $400$ and $7.65$, respectively, for all methods (the range of image values is $[0\thicksim255]$).

For convergence analysis of the WLS smoothing, we compare our approach (FWLS) with the conjugate gradient (CG), Gauss-Seidel (GS), Jacobi iteration (JI), and successive over relaxation (SOR). We use the SuiteSparse library \cite{suitesparse} to solve a triangular system, arising from GS and SOR.
In the case of the WTV smoothing, our approach (FWTV) is compared to the split Bregman method (SB) \cite{Bi2015,Goldstein2009} and classical penalty decomposition (classical-PD) \cite{Wang2008}.
The FFTW library \cite{FFTW} is used to solve the $u$-subproblem of (5) for SB and PD-classical methods.
We show in Fig. 2(a) how the WLS objective evolves at each iteration (all methods are initialized by the input $f$).
Although each iteration of CG, GS, JI, and SOR runs in a linear time, they require a very large number of iterations to converge.
In contrast, our solver converges in a few iterations.
The result of the WTV is shown in Fig. 2(b).
Likewise, our approach converges much faster than SB and classical-PD.
Note that these methods have $O(n\log n)$ complexity per iteration $(n=H\times W)$, while our solver runs in linear time.
The input image is shown in the inset Fig 1.

\begin{figure*}[t]
\centering
\renewcommand{\thesubfigure}{}
\subfigure[(a) The WLS smoothing]
{\includegraphics[width=0.49\linewidth]{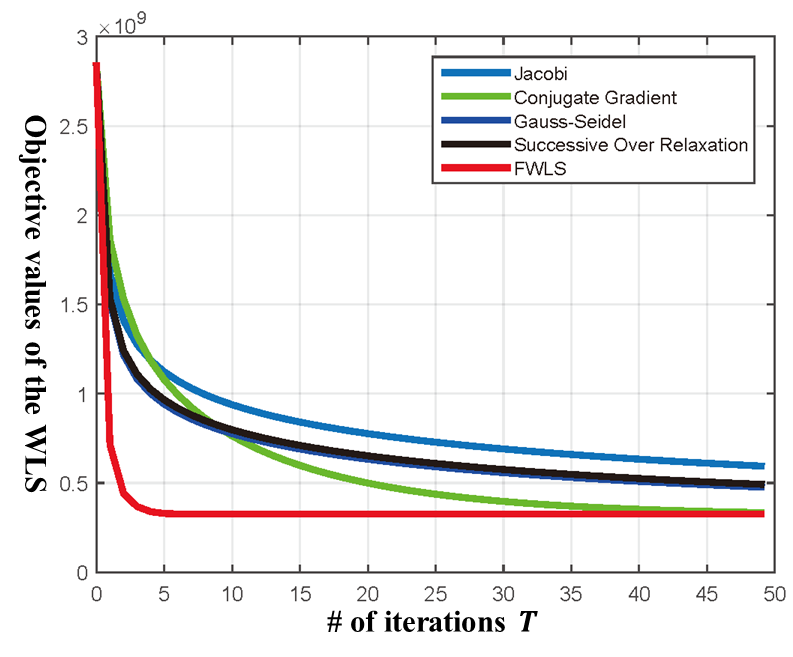}}\hfill
\subfigure[(b) The WTV smoothing]
{\includegraphics[width=0.492\linewidth]{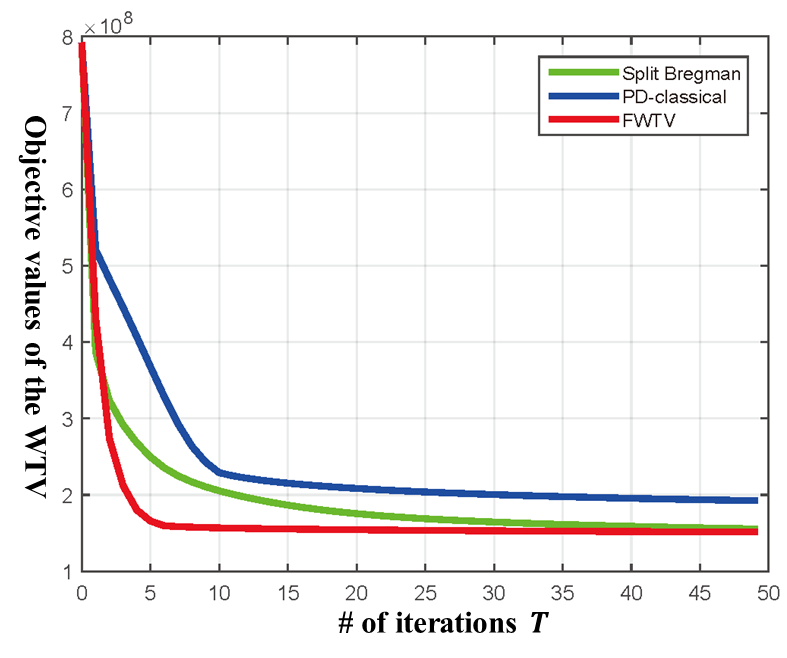}}\hfill
\caption{Comparison of the objective decrease versus the number of iterations. (a) the WLS smoothing and (b) the WTV smoothing.
Our approach rapidly reduces the objective value of (1), and converges after a few iterations (the input image is shown in the inset of Fig. 1). $\beta^{1}$ and $\alpha$ are set to $1$ and $4$, respectively.}
\label{img:2}
\end{figure*}

\begin{figure*}[t]
\centering
\renewcommand{\thesubfigure}{}
\subfigure[]
{\includegraphics[width=1\linewidth]{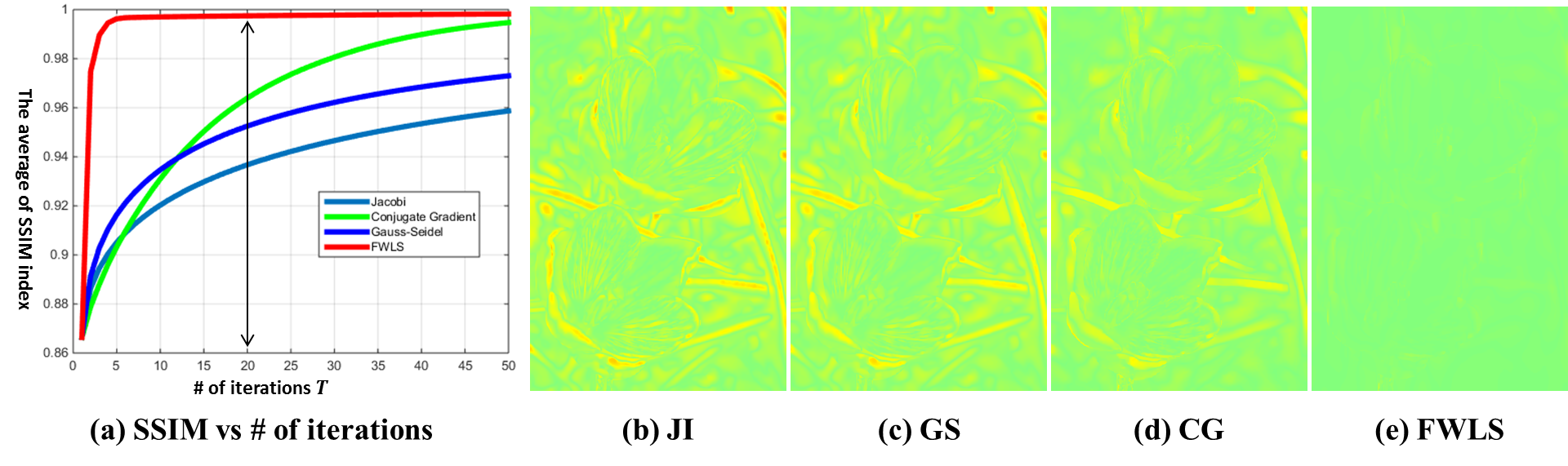}}\hfill
\vspace{-15pt}
\caption{The average SSIM indexes \cite{Wang2004} for the WTV smoothing, as a function of the number of iterations. (a) The average SSIM values on 100 natural images,
(b)-(d) The visualization of differences between the reference and the results of each method (at 20 iteration).}
\label{img:2}
\end{figure*}

\begin{table}[]
\centering
\caption{Runtime comparison for different methods (in seconds)}
\label{my-label}
\begin{tabular}{|c|
>{\columncolor[HTML]{EFEFEF}}c |c|c|c|
>{\columncolor[HTML]{EFEFEF}}c |c|c|c|}
\hline
Image size & \cellcolor[HTML]{EFEFEF}                      & PCG \cite{Krishnan2013} & MATLAB ``$\;\backslash$ " & FWLS & \cellcolor[HTML]{EFEFEF}                      & SB \cite{Bi2015,Goldstein2009} & FWTV \\ \cline{1-1} \cline{3-5} \cline{7-8}
427x640    & \cellcolor[HTML]{EFEFEF}                      & 0.85   & 0.95                      & 0.07 & \cellcolor[HTML]{EFEFEF}                      & 1.47  & 0.25 \\ \cline{1-1} \cline{3-5}\cline{7-8}
660x800    & \cellcolor[HTML]{EFEFEF}                      & 1.58   & 1.97                      & 0.13 & \cellcolor[HTML]{EFEFEF}                      & 2.56  & 0.43 \\ \cline{1-1} \cline{3-5}\cline{7-8}
923x1128   & \multirow{-4}{*}{\cellcolor[HTML]{EFEFEF}WLS} & 3.04   & 4.14                      & 0.28 & \multirow{-4}{*}{\cellcolor[HTML]{EFEFEF}WTV} & 6.69  & 1.02 \\ \hline
\end{tabular}
\end{table}

To further demonstrate the effectiveness of our approach, we additionally collect 100 natural images from the BSDS500 dataset \cite{Arbelaez11}, and compare the WLS smoothing results.
The smoothing result obtained by MATLAB ``$\;\backslash$ " command is used as the reference.
The average value of the structural similarity (SSIM) index \cite{Wang2004} is plotted as a function of the number of iterations in Fig. 3(a).
The difference images between the reference and the results at 20 iteration is also visualized in Fig. 3(b)-(e).
In general, we find the proposed approach yields satisfactory results after $T=3\sim5$.
The average SSIM values at $T=3$, $5$, and $20$ are 0.9896, 0.9963, and 0.9975, respectively.

The runtime comparison is shown in Table 1.
In our methods, the number of iterations $T$ is fixed to 5 based on the above experiment.
For the WLS smoothing, our approach is compared with the state-of-the-art preconditioning method (PCG) \cite{Krishnan2013} and MATLAB ``$\;\backslash$ " operator.
We note that MATLAB ``$\;\backslash$ " operator uses the sparse cholesky decomposition in the SuiteSparse library \cite{suitesparse}.
The result for the PCG \cite{Krishnan2013} is obtained from source code provided by the author.
It should be noted that although the preconditioning method \cite{Krishnan2013} improves the rate of convergence significantly, constructing the preconditioner needs lots of setup time, taking about 2.6 second for 1$M$ image.
The stopping criteria of the SB \cite{Bi2015,Goldstein2009} is ${\left\| {{u^{k + 1}} - {u^k}} \right\|_2}<0.1$.
Our approach is magnitude faster than competing methods.

Next, we present the analysis of our fast iteratively re-weighted algorithms.
Using the FIRLS, we first minimize the objective\footnote{The same image in Fig. 1 is used and $\sigma$ is set to 7.65} (17) with $\psi  = \sigma (1 - \exp ( - \frac{{{\tau^2}}}{\sigma }))$.
Fig. 4(a) shows that the convergence rate of the FIRLS method differs depending on the number of inner iterations $T$.
For the FIRL1, we use $\psi  = \log (1 + |\tau|)$ as it is not well suited for the potential function that has a quadratic behavior around 0 (Fig. 4(b)).
Overall, we observe that the iteratively re-weighted algorithms are stuck in bad local minima if internal iterations are not carried out sufficiently.
The objective value even fluctuates with $T=1$, as the red line of Fig. 4(a).
It is thus crucial to solve the subproblems of iteratively re-weighted algorithms until reaching the certain level of accuracy.
In general, we find $K=5$, $T=5$ is a good choice for both algorithms\footnote{On this example, an average value of per-pixel difference, i.e., ${\left\| {{u^{k + 1}} - {u^k}} \right\|_2}$ is 0.15 after $K=5$ external iterations (the range of image values is $[0\thicksim255]$).}.
Many heavy-tailed potential functions $\psi$ are non-convex, and thus any warm initializations for $u^1$ may further improve the convergence of our fast iteratively re-weighted algorithms.

\begin{figure*}[t]
\centering
\renewcommand{\thesubfigure}{}
\subfigure[(a) FIRLS $\psi  = \sigma (1 - \exp ( - \frac{{{\tau^2}}}{\sigma }))$]
{\includegraphics[width=0.49\linewidth]{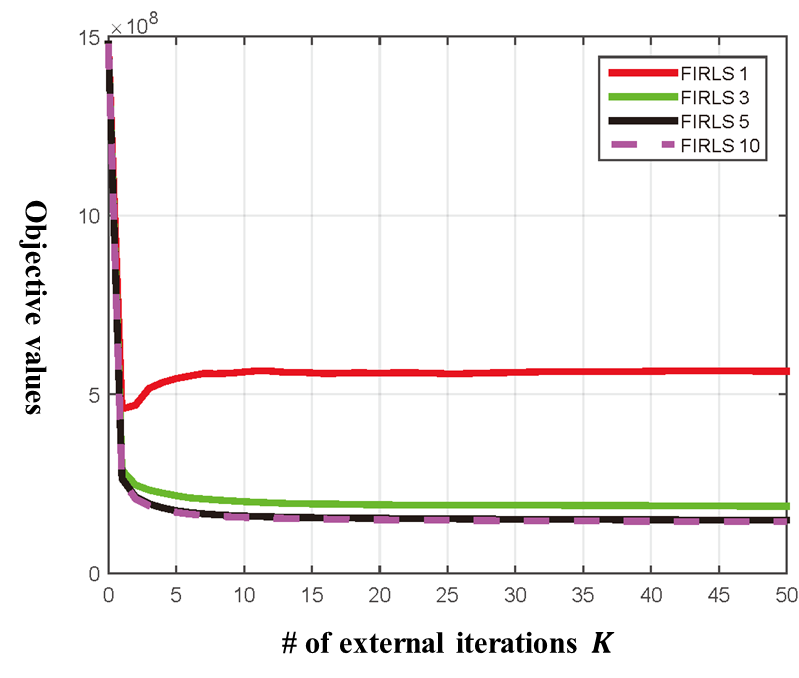}}\hfill
\subfigure[(b) FIRL1 $\psi  = \log (1 + |\tau|)$]
{\includegraphics[width=0.49\linewidth]{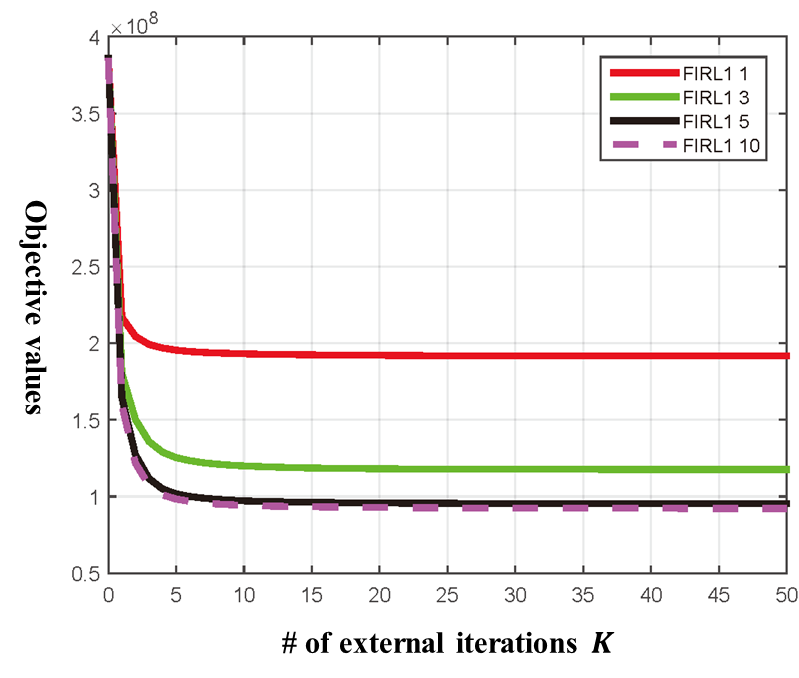}}\hfill
\caption{Convergence of iteratively re-weighted algorithms depending on the number of inner iterations $T$ (please see legend). (a) the FIRLS and (b) the FIRL1. Each convex surrogate functions are solved with our FWLS and FWTV solvers. If each internal iteration is not performed sufficiently, the iteratively re-weighted algorithms stuck in bad local minima.}
\label{img:2}
\end{figure*}

\begin{figure*}[t]
\centering
\renewcommand{\thesubfigure}{}
\subfigure[]
{\includegraphics[width=1\linewidth]{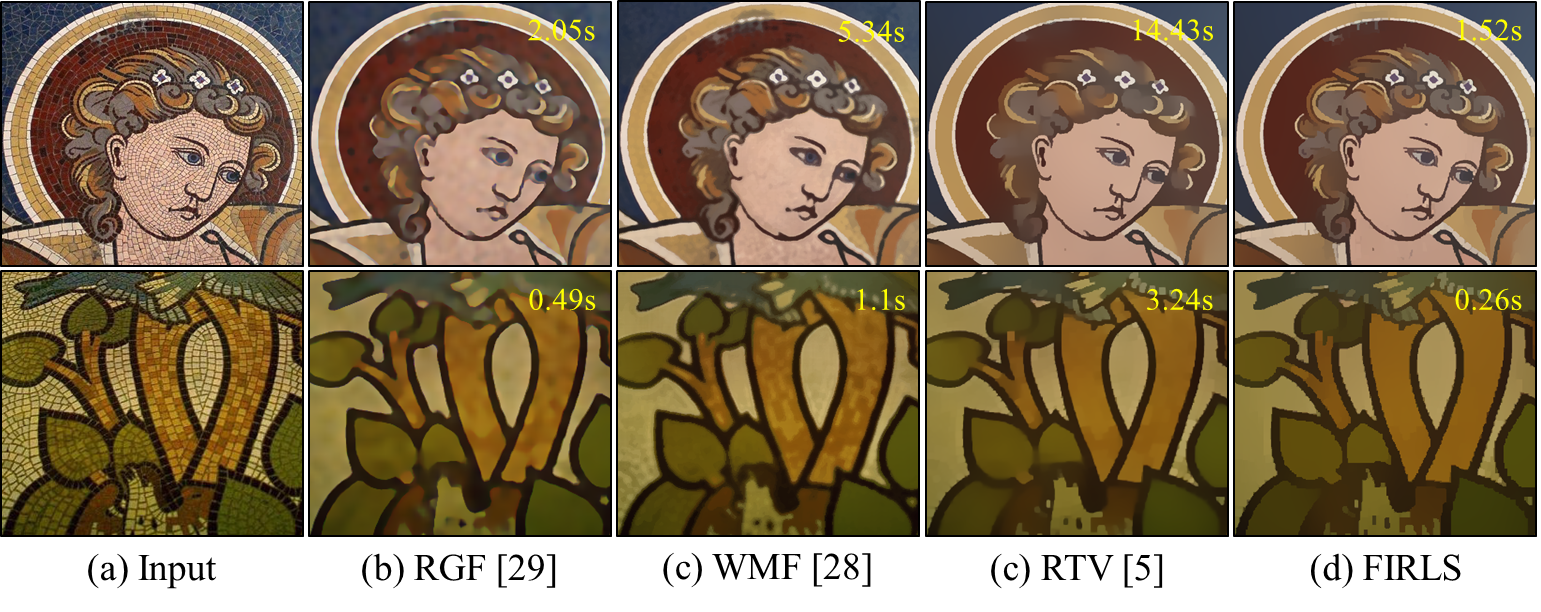}}\hfill
\vspace{-15pt}
\caption{Examples of the texture removal.
For each method, the running times are reported in seconds (yellow). The image sizes are $1254\times 1067$ (top) and $495\times 536$ (bottom), respectively.}
\label{img:2}
\end{figure*}

\begin{figure*}[t]
\centering
\renewcommand{\thesubfigure}{}
\subfigure[]
{\includegraphics[width=1\linewidth]{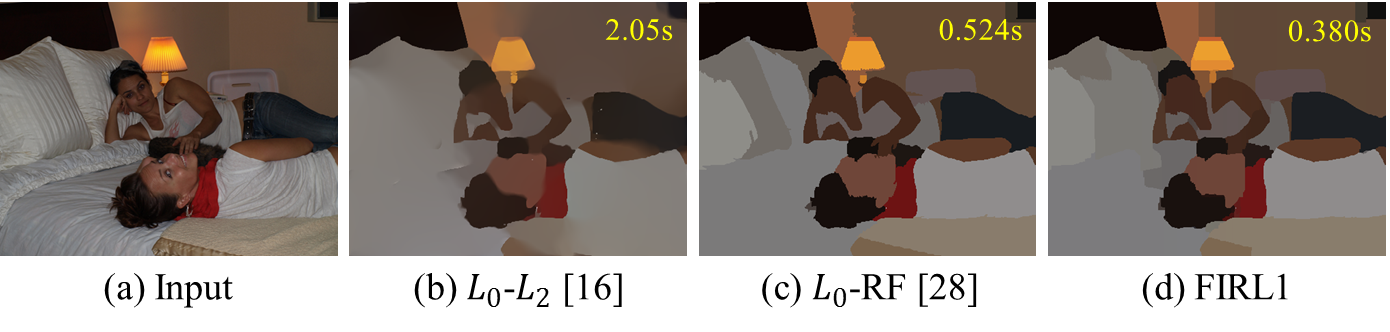}}\hfill
\vspace{-20pt}
\caption{Examples of the content-based color quantization. Our approach shows performance comparable to state-of-the-art $L_0$ minimization \cite{Nguyen2015}.
The running time is also reported in seconds (the image sizes are $494\times 371$).}
\label{img:2}
\vspace{-10pt}
\end{figure*}

\begin{figure*}[t]
\centering
\renewcommand{\thesubfigure}{}
\subfigure[]
{\includegraphics[width=1\linewidth]{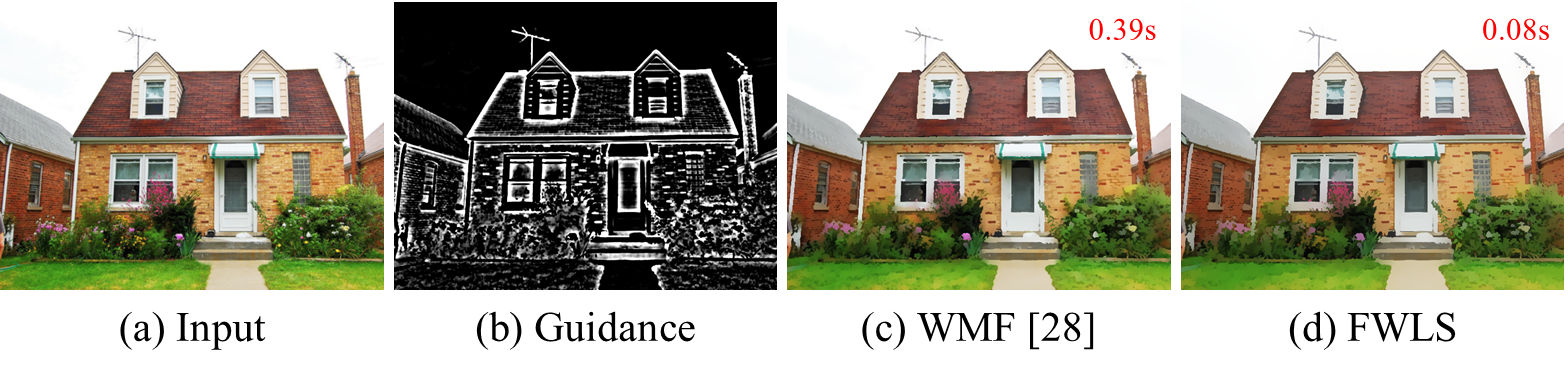}}\hfill
\vspace{-20pt}
\caption{Style transfer effect. The edge map of the input is used as the guidance image. The image size is $640\times 480$. The input image (a) is taken from \cite{Zhang20142}.}
\label{img:2}
\end{figure*}

\section{Applications}
Our approach is flexible, and can be applied to several tasks using EPS. In this section, we apply our method to texture removal, content-based color quantization, and style transfer.
To this end, the various image priors, i.e., potential function and weight $w$, are exploited to match different application goals.
All results and runtime in the comparison are obtained from source codes provided by the authors.
The parameters are carefully tuned through extensive experiments.
Additional results and other applications, such as scale-space filtering and image denoising, are also available in the supplementary material.

\subsection{Texture removal}
For texture removal, we employ the model proposed in \cite{Ham2015}: $\psi$ is set to $\sigma (1 - \exp ( - {{{\tau^2}}}/{\sigma }))$ and $g=G*f$ where $G$ is the Gaussian kernel with standard deviation 2.
This type of guidance image is very effective since textures on the object are usually of small scale structures.
The smoothing parameters $\kappa$, and $\sigma$ are fixed to 5 and 7.65, respectively, but $\lambda$ varies according to images.
In this setting, we minimize the objective (17) using our FIRLS solver.
Fig. 5 shows examples of texture removal obtained by the rolling guidance filter (RGF) \cite{Zhang2014}, the weighted median filter (WMF) \cite{Zhang20142},
the relative total variation (RTV) \cite{Xu2012}, and our FIRLS. The RGF \cite{Zhang2014} is implemented using fast bilateral filter \cite{Chen2007}.
The proposed method is even faster than the texture smoothing tools based on the fast local filtering (RGF \cite{Zhang2014}, WMF \cite{Zhang20142}), while outperforming them in the subjective evaluation.
Note that minimizing the objective function in the RTV \cite{Xu2012} needs to solve a large linear system iteratively\footnote{In the RTV \cite{Xu2012}, the prior term is defined in a form of total variation, resulting in a nonlinear system of equation. It can be easily addressed by using a fixed point iteration.}. Thus, it can be also accelerated by our FWLS solver.

\subsection{Color quantization}
(Content-based) color quantization attempts to increase the sparsity of colors while maintaining the overall structure in images.
This is useful for many vision tasks, including image retrieval and segmentation.
We use the sparsity prior, $\psi  = \log (1 + |\tau|)$, to reduce the number of colors.
The guidance image is not used in this application, i.e., $w=1$.
With this, we minimize the objective (17) using our FIRL1 solver.
Results and comparisons are provided in Fig. 6.
Our solver shows performance comparable to state-of-the-art $L_0$ minimization \cite{Xu2011,Nguyen2015}, while running faster.
The result in Fig. 6(c) has been obtained using the region fusion (RF) approach \cite{Nguyen2015}, which is tailored to $L_0$ minimization only.

\subsection{Style transfer}
Any feature of input image or others can be taken as the guidance $g$.
To transfer structure of $g$ to input images $f$, we use our FWLS and FWTV solvers.
Fig. 7 shows a specific style transfer example obtained by the weighted median filter (WMF) \cite{Zhang20142} and our FWLS.
The edge map of $f$ is used as the guidance image $g$. The window size of the WMF \cite{Zhang20142} is set to $5\times5$, taking about 0.4 seconds.
Our FWLS and FWTV solvers take only 0.08 and 0.28 seconds, respectively, for an image of size $640\times 480$.

\section{Conclusions}
We introduce a highly efficient splitting-based method for global EPS.
Unlike previous splitting-based methods, our formulation enables linear time solvers for
WLS and WTV problems.
Our solver converges quickly, and its runtime is comparable to state-of-the-art local EPS approaches.
We also propose fast iteratively re-weighted algorithms for a non-convex objective function.
Our approach is flexible, and thus is applicable to a variety of applications.
\section{Acknowledgements}
\pagebreak

\bibliographystyle{splncs}
\bibliography{egbib}
\end{document}